\documentclass{article}

\usepackage{arxiv}

\usepackage[utf8]{inputenc} % allow utf-8 input
\usepackage[T1]{fontenc}    % use 8-bit T1 fonts
\usepackage{hyperref}       % hyperlinks
\usepackage{url}            % simple URL typesetting
\usepackage{booktabs}       % professional-quality tables
\usepackage{amsfonts}       % blackboard math symbols
\usepackage{nicefrac}       % compact symbols for 1/2, etc.
\usepackage{microtype}      % microtypography
\usepackage{lipsum}		% Can be removed after putting your text content
\usepackage{graphicx}
\usepackage{natbib}
\usepackage{doi}

\usepackage{amsmath,amssymb}
\usepackage{caption}

\usepackage{enumerate}
\usepackage{multirow}

\usepackage{colortbl}

\title{Early Prediction of Natural Gas Pipeline Leaks Using the MKTCN Model}

\author{ \href{https://orcid.org/0009-0001-0385-2925}{\includegraphics[scale=0.06]{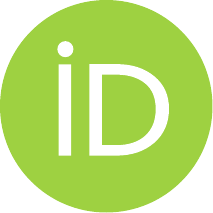}\hspace{1mm}Xuguang Li}\\
	School of Electrical and Automation\\
	Wuhu Institute of Technology\\
	Wuhu, China \\
	\texttt{101146@whit.edu.cn} \\
	\And
	\href{https://orcid.org/0000-0001-6283-1668}{\includegraphics[scale=0.06]{orcid.pdf}\hspace{1mm}Zhonglin Zuo} \\
	School of Control Science and Engineering\\
	Zhejiang University\\
	Hangzhou, China \\
	\texttt{z.zuo@zju.edu.cn} \\
        \And
	\href{https://orcid.org/0009-0001-7055-7766}{\includegraphics[scale=0.06]{orcid.pdf}\hspace{1mm}Zheng Dong} \thanks{Corresponding author: Zheng Dong.} \\
	ByteDance\\
	Beijing, China \\
	\texttt{dongzheng@bytedance.com} \\
         \And
	\href{https://orcid.org/0000-0002-5245-3584}{\includegraphics[scale=0.06]{orcid.pdf}\hspace{1mm}Yang Yang}\\
	School of Computer Science and Engineering\\
        Nanjing University of Science and Technology\\
	Nanjing, China \\
	\texttt{yyang@njust.edu.cn} \\
}

\hypersetup{
pdftitle={Early Prediction of Natural Gas Pipeline Leaks Using the MKTCN Model},
pdfsubject={CoRR.AI, eess.SP},
pdfauthor={Xuguang Li, Zhonglin Zuo, Zheng Dong},
pdfkeywords={Natural gas pipeline leaks, Early prediction, Multi-dimensional time-series, MKTCN model, Sample imbalance, Long-term dependencies},
}

\begin{document}
\maketitle

\begin{abstract}
Natural gas pipeline leaks pose severe risks, leading to substantial economic losses and potential hazards to human safety. In this study, we develop an accurate model for the early prediction of pipeline leaks. To the best of our knowledge, unlike previous anomaly detection, this is the first application to use internal pipeline data for early prediction of leaks. The modeling process addresses two main challenges: long-term dependencies and sample imbalance. First, we introduce a dilated convolution-based prediction model to capture long-term dependencies, as dilated convolution expands the model's receptive field without added computational cost. Second, to mitigate sample imbalance, we propose the MKTCN model, which incorporates the Kolmogorov-Arnold Network as the fully connected layer in a dilated convolution model, enhancing network generalization. Finally, we validate the MKTCN model through extensive experiments on two real-world datasets. Results demonstrate that MKTCN outperforms in generalization and classification, particularly under severe data imbalance, and effectively predicts leaks up to 5000 seconds in advance. Overall, the MKTCN model represents a significant advancement in early pipeline leak prediction, providing robust generalization and improved modeling of the long-term dependencies inherent in multi-dimensional time-series data.
\end{abstract}

% keywords can be removed

\section{Introduction}
Natural gas pipeline leaks pose significant safety hazards and can lead to considerable economic losses \citep{psyrras2018safety}. The current research on pipeline leakage is primarily concerned with detecting fatigue \citep{waijie}, \citep{jiqiren1} and leakage anomalies \citep{pipedetect5}, \citep{0Leak}, \citep{zuo2022semi}. These methods are highly accurate. However, pipeline fatigue detection methods necessitate direct pipeline manipulation and are inapplicable to long-distance pipelines. Leakage anomaly detection methods diagnose operating conditions by analyzing the physical properties of the transport medium, such as temperature and pressure, which allows them to bypass the need for direct pipeline operation. Nevertheless, this method remains an after-the-fact detection method. Therefore, it is urgent to establish an accurate and effective leakage prediction method for long-distance pipelines.

Early prediction is a widely applied technique in healthcare \citep{medicine2}, industry \citep{predictIndustry2}, and finance \citep{bluwstein2023credit}. However, a large amount of data is required to build predictive models for machine learning \citep{dean2014big}, \citep{shen2021topic}, \citep{hang2022outside}, \citep{ye2022mane}, \citep{yang2023contextualized}. Supervisory Control and Data Acquisition (SCADA) systems are widely employed in the natural gas pipeline industry \citep{rao2017industrial}, where vast amounts of data concerning transported substances are collected and stored. These datasets provide direct insights into the pipeline's operational conditions and are inherently time-series. By the Bernoulli equation for fluid flow in a pipe, it is evident that the flow rate, pressure, and other characteristics at various points within the fluid, along with the shape and height of the pipe, are closely interconnected \citep{liu2014social}, \citep{harriott2011gas}. Even slight deformations in the pipeline's structure can lead to changes in the motion of the transported material \citep{de2021unstable}. In the early stages of a natural gas pipeline leak, specific internal signals, such as abnormal pressure fluctuations and slight changes in flow rate, begin to emerge \citep{quy2021real}. These subtle indicators offer valuable information that can be leveraged to predict leaks at an early stage. As natural gas pipelines are long-distance, this manifests itself in leakage data that has a long-term dependence. In other words, data exhibiting a high degree of time dependence offer excellent support for the early prediction of leakage. Therefore, developing an accurate mathematical model of the pipeline based on this data would be highly beneficial. However, the infrequent occurrence of anomalies, coupled with the numerous factors that can cause them. The need for more abnormal data leads to a highly imbalanced sample set, which fails to meet the essential requirements for building a robust mathematical model \citep{zheng2017querying}, \citep{dong2021butterfly}. While Generative Adversarial Networks (GANs) \citep{2014Generative} can mimic original data distributions, our research finds that TimeGAN \citep{NEURIPS2019_c9efe5f2} underperforms in real-world data generation. Oversampling \citep{2002SMOTE} and undersampling \citep{liu2008exploratory} address sample imbalance, but oversampling may distort real scenarios, and undersampling yields suboptimal results with scarce anomalous data. Therefore, when modeling the early prediction of pipeline leakage, we need to address the challenges of sample imbalance and the long-term dependence of the time series.

This paper examines the challenges associated with long-term data dependency and sample imbalance in the context of early prediction of natural gas pipelines. To this end, we propose a novel model construction approach. It is observed that the dataset exhibits an intrinsic characteristic of being potentially leaky in the period preceding the actual leak, which we term as \textit{doubtful}. Firstly, we model the time series early prediction task as a multi-class classification task, allowing the model to accurately classify \textit{doubtful} data to indicate its predictive capabilities. Secondly, we devised a method to restructure the original time series data into serial format, enhancing the model's computational efficiency and generalization ability. Finally, we propose the Multi-classification Temporal Convolutional Network with the Kolmogorov-Arnold Network (MKTCN), which tackles the challenge of long-term dependence in pipeline data by utilizing dilated convolution. Furthermore, the MKTCN replaces the fully connected layer with the Kolmogorov-Arnold Network (KAN) \citep{liu2024kan}, effectively mitigating the issue of sample imbalance. The contributions of this paper are summarized as follows:

\begin{enumerate}[1)]

\item {First Application of Early Predictive Modeling for Pipeline Leaks}: To our knowledge, this study represents the first application of internal pipeline signals for the early prediction of gas pipeline leaks. Unlike previous studies focusing on detecting anomalies post-leak, our approach aims to anticipate leaks in advance.

\item {Addressing the Challenge of Long-Term Dependence on Pipeline Data}: Due to the extensive length of natural gas pipelines, data transmission exhibits significant long-term dependencies. Therefore, we developed an early prediction model for pipeline leakage based on the dilated convolutional model to solve the long-term dependencies problem. The dilated convolutional model offers a broader perspective than the traditional model while requiring no additional computational resources.

\item {Addressing the Challenge of Imbalanced Pipeline Leakage Data Samples }: We use the KAN network to address the challenge of sample imbalance encountered during the early prediction of pipeline leaks. Given the low probability of leak events, we extend the traditional TCN model by replacing its fully connected layer with a KAN. Overall, we innovatively propose a novel MKTCN model to solve the task of early prediction of pipeline leakage.

\end{enumerate}

The rest of this paper is organized as follows. We describe in Section \ref{sec: METHODOLOGY} how we define the gas pipeline early prediction problem and the architecture of the MKTCN model used to solve it. Section \ref{sec: Experiments} shows the experiments and corresponding results. Finally, Section \ref{sec: Conclusion} summarizes and concludes this paper.

\section{Proposed framework}
\label{sec: METHODOLOGY}
This section presents the methodology employed in defining the early prediction problem for natural gas pipelines and accounts for the ideas and methods used to solve it.

\subsection{Problem statement}
\label{sec:Problem Statement}

\begin{figure*}[t]
\centering 
\includegraphics[width=6.5in]{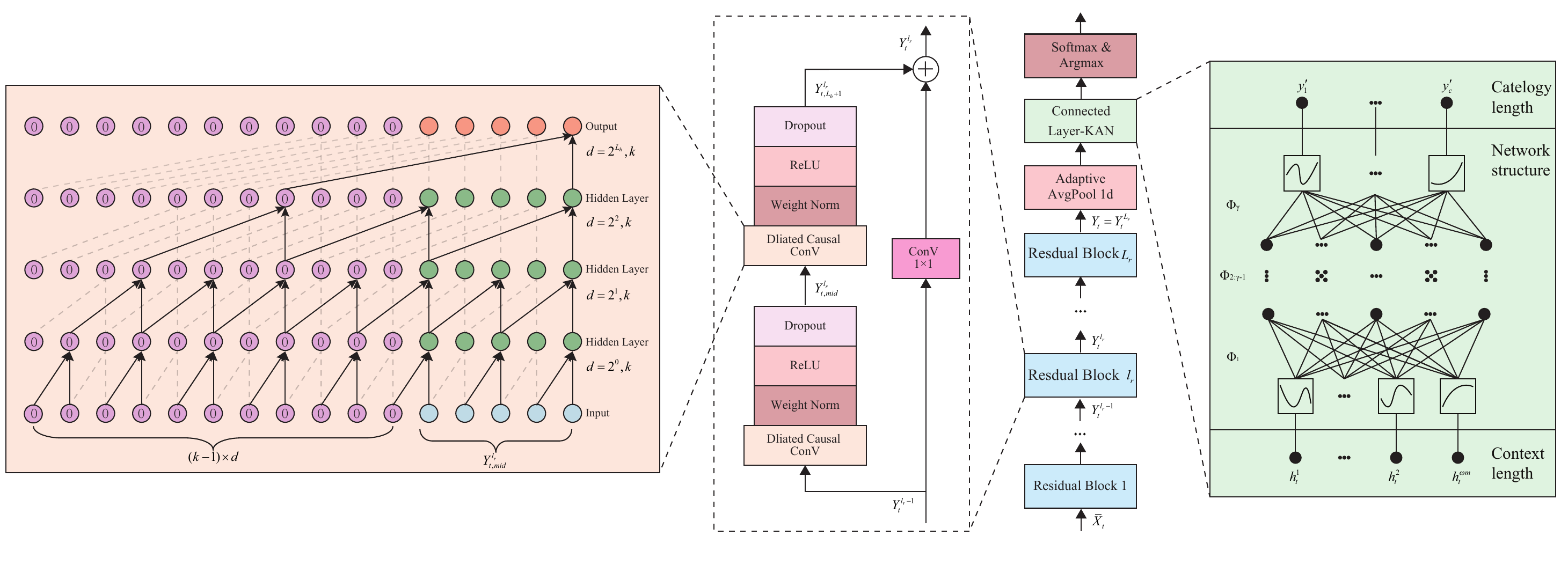} 
\caption{The overall architecture of the MKTCN model.} 
\label{fig:model} 
\end{figure*}

\begin{figure}[t]
\centering 
\includegraphics[width=2.5in]{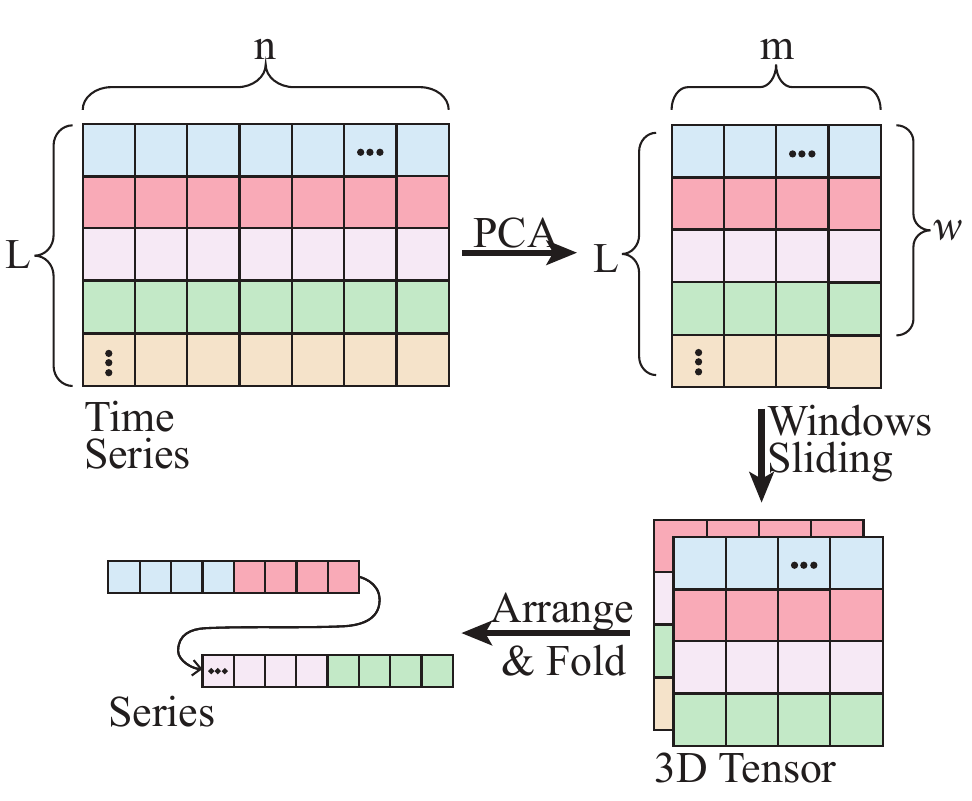} 
\caption{Data processing method. Using this method, we transformer time-series data into serial data, improving generalization and removing the long-term dependence on the data.} 
\label{fig: data process} 
\end{figure}

The sensors operate in conjunction when the SCADA system is operational. The reading of the $i$-th sensor at the moment $t$ is recorded as $x_{t}^{i}$. It is not easy to fully describe the operational state of the pipeline using only a single sensor. Therefore, collecting readings from as many sensors along the pipeline is feasible. The readings from all sensors at time stamp $t$ can be denoted as $X_{t}=\left\{ x_{t}^{1},x_{t}^{2},\dots,x_{t}^{n} \right \} \in \mathbb{R}^{n}$, where $n$ is the total number of sensors. The SCADA system can use the anomaly detection model \citep{pipedetect5}, \citep{zuo2022semi} to analyze the pipeline leakage in real-time and label the \textit{normal} data $y=0$ and label \textit{abnormal} data $y=1$. If a leak occurs in the pipeline at timestamp $t_{e}$, then the data within the $N$ time units preceding the leak is considered potentially indicative of the leak. In other words, the label the \textit{doubtful} data $\left\{y_{e-1}, y_{e-2},\cdots,y_{e-N}\right\}$ $y=2$. Therefore, we have the dataset $\mathcal{H}_{T}=\left\{ \mathcal{X}_{T},Y_{T}\right \}=\left\{(X_0,y_0),(X_1,y_1),\dots,(X_{T-1},y_{T-1})\right \} \in \mathbb{R}^{(n+1) \times T}$, where $T$ is the total number of observations and $y \in \left\{0,1,2\right \}$.

It allows us to solve the early prediction task with a multi-classification model. Constructing a multi-classification task model enables early leakage prediction, and the model's classification ability is reflected in its prediction accuracy.

Early prediction aims to provide a warning before anomalies become evident. We introduce a time-series data processing technique that not only preserves the inherent characteristics of the original dataset but also mitigates long-term temporal dependencies. We model the early anomaly prediction task as a multi-class classification problem, significantly enhancing the model's ability to detect early warning signals of potential leaks.

\subsection{MKTCN}

This paper proposes a novel model, MKTCN, which combines Temporal Convolutional Networks (TCN) \citep{baseline-tcn} and KAN to address temporal dependencies and sample imbalance in series multi-classification tasks. The overall architecture of the model is shown in Fig. \ref{fig:model}. The MKTCN employs dilated convolutions to capture long-term dependencies in time series data, enabling the model to utilize information over extended time spans while maintaining computational efficiency. Concurrently, the KAN, founded on the Kolmogorov-Arnold representation theorem, improves the model's generalization ability when dealing with imbalanced data distributions, especially excelling in classifying minority classes. By integrating the strengths of both networks, the MKTCN model not only effectively addresses the issue of long-term dependencies in sequential data but also enhances classification performance on imbalanced datasets.

To improve the data quality, we must pre-process the original time series data $\mathcal{X}_{T}$ before building the model. Fig. \ref{fig: data process} illustrates the data processing steps. The first step for original labeled time series data involves using Principal Component Analysis (PCA) \citep{turk1991eigenfaces} to reduce dimensionality. On the one hand, PCA helps eliminate noise and redundant information, making subsequent analysis more efficient. On the other hand, dimensionality reduction improves the model's generalization capacity and reduces the risk of overfitting. The PCA-processed data is $\mathcal{X}_{T}^{\prime}=\left\{X_0^{\prime},X_1^{\prime},\dots,X_{T-1}^{\prime}\right \} \in \mathbb{R}^{m \times T}$, where $\mathcal{X}_{T}^{\prime}$ still retains the time series format, $m$ is the dimension after dimensionality reduction and $m \leq n$. Next, $\mathcal{X}_{T}^{\prime}$ is transformed into a 3D tensor format using a sliding window of length $\omega$, and the number of tensor data is $\lfloor \frac{T-\omega}{s} \rfloor +1$, where $s$ is the sliding step of windows and the symbol $ \lfloor * \rfloor$ is floor function. Each tensor data is labeled with the latest timestamp it contains. Finally, each tensor is structured and folded into a data sequence $\Bar{{X}}_{T} \in \mathbb{R}^{\omega m}$, with each sequence assigned the label of its corresponding tensor. This approach helps decorrelate the temporal dependencies in the time series data, further enhancing the model's generalization ability \citep{zhang2024smde}. 

As a result of this data processing, the initial anomaly detection dataset $\mathcal{X}_{T}$ was transformed into a one-dimensional classification dataset $\Bar{\mathcal{X}}_{T} $. In addition to improving data quality, this process converts the early anomaly prediction task into a multi-class classification problem, which boosts the model's ability to generalize.

To give our model the ability to model long-time data, we use dilation convolution to address it. As illustrated in Fig. \ref{fig:model}, the MKTCN model has been designed with $L_r$ residual blocks, each comprising two identical dilation convolution modules. Dilation convolutions with $L_h$ hidden layers primarily constitute the dilation convolution module. Therefore, at moment $t$, the $l_n$-th neuron of the $l_h$-th hidden layer for the $l_r$-th residual block is represented as follows:
\begin{equation}
\begin{aligned}
\label{eq: y}
y_{t,l_h}^{l_r,l_n}=\sum_{i=0}^{k}({w}_{i} \cdot y_{t,l_h-1}^{l_r,l_n-di}+{q}^{l_h})
\end{aligned}
\end{equation}  
where $0 \leq l_n \leq \omega m$, $0 \leq l_h \leq L_h$, $0 \leq l_r \leq L_r$, $k$ is the size of the convolution kernel, and $d=2^{l_h}$ is the dilatation factor. Meanwhile, $w_{i}$ and $q^{l_h}$ are the learnable parameters. 

Therefore, the output of the $l_h$-th hidden layer can be represented as $Y_{t,l_h}^{l_r}=\{ y_{t,l_h}^{l_r,1},y_{t,l_h}^{l_r,2},\cdots,y_{t,l_h}^{l_r,\omega  m} \}$. It is important to note that $Y_{t,L_h+1}^{l_r}$ results from the output layer. Dilated convolution is a principal component of the MKTCN model, which enables the model to have a more expansive receptive field, thereby facilitating the capture of longer sequence dependencies. Dilated convolution confers upon the MKTCN model a superior capacity to process sequence data. 

It is necessary to apply a data padding procedure before the model process to prevent the inadvertent disclosure of data information. It involves inserting an additional 0 at the start of the series, ensuring that the model is only analyzed using data from before the current moment. It is noteworthy that the additional 0 does not increase the computation. The padding operation can be represented as follows:
\begin{equation}
\begin{aligned}
\label{eq: pad series}
\Bar{Y}_{t,l_n}^{l_r}=\{\underbrace{0, \dots, 0}_{(k-1) \times d}, Y_{t,l_n}^{l_r}\}  \in \mathbb{R}^{\omega  m + (k-1) d}
\end{aligned}
\end{equation} 
therefore, the input of the first hidden layer can be represented as $Y_{0,0}^{0}=\{0,\cdots,0,\Bar{X}_{0}\} \in \mathbb{R}^{\omega m + k -1}$ and the padding operation ensures that $y_{t,l_n-1}^{l_r,l_n-di}=0$ in Eq.\ref{eq: y} when $l_n-di<0$. The padding operation keeps the length of the output sequence consistent with the input sequence's length and prevents data leakage. It is important to note that implementing causal convolution depends on the padding operation. 

To build an effective sequence model that captures long-range dependencies, each residual block is made up of two dilated convolutional networks with the same parameters \citep{baseline-tcn}. The weight-norm and dropout layers are added to each layer to regularize the network. Concurrently, each residual block employs residual connections to incorporate inputs into the output of the convolutional layer, thereby facilitating jump connections. Therefore, the $l_r$-th residual connection can be expressed as follows:
\begin{equation}
\begin{aligned}
\label{eq:residual connections}
Y_{t,L_h+1}^{l_r}&=\sigma\left(\mathcal{W}^{l_r} \cdot Y_{t,mid}^{l_r}+Q^{l_r}\right)\\
Y_{t,mid}^{l_r}&=\sigma\left(\mathcal{W}^{l_r} \cdot Y_{t}^{l_r-1}+Q^{l_r}\right)
\end{aligned}
\end{equation} 
where $\sigma(*)$ is the non-linear activation function, $\mathcal{W}^{l_r}$, $Q^{l_r}$ are the learnable parameters of the $l_r$-th layer of the dilation convolution, respectively.

Meanwhile, the MKTCN model uses a convolutional network of size $1 \times 1$ to connect the input and the output sequence as follows:
\begin{equation}
\begin{aligned}
\label{eq: convolutional network 1*1}
Y_{t}^{l_r}=Y_{t,L_h+1}^{l_r}+Y_{t}^{l_r-1}
\end{aligned}
\end{equation}

Therefore, in the MKTCN network comprising $L_r$ residual block, the final output of the model is as follows:
\begin{equation}
\begin{aligned}
\label{eq:final output}
Y_{t}=Y_{t}^{L_r}=\{h_{t}^{1},h_{t}^{2},\dots ,h_{t}^{\omega m}\}
\end{aligned}
\end{equation} 

To address the issue of sample imbalance, the KAN model is used to create a connection between the output of residual block layer $Y_{t}$. Based on Kolmogorov-Arnold representation theorem \citep{liu2024kan}, a generic deeper KAN layer can be expressed by the composition $\gamma$ layers as follows:
\begin{equation}
\begin{aligned}
\label{eq:kan}
H_t=(\Phi_{\gamma}\circ \Phi_{\gamma-1} \circ \dots \circ \Phi_{1}) \cdot Y_t
\end{aligned}
\end{equation} 
where $\circ$ is combinatorial operation of functions, and $\Phi_{l}$ is a KAN activation function of $l$-th layer. Meanwhile, the activate matrix $\Phi_{l}$ composed by univariate function $\{ \phi_{q^l,p^{l}} \}$ with $p^{l}=1,2,\dots,n_{in}^{l}$ and $q^{l}=1,2,\dots,n_{out}^{l}$, where $n_{in}^{l}$ and $n_{out}^{l}$ are the number of inputs and outputs of the $l$-th layer activation function, respectively. It is noteworthy that $n_{in}^{0}=\omega  m$ and $n_{out}^{\gamma}=c$.

In particular, we can describe the calculation process as follows:
\begin{equation}
\begin{aligned}
\label{eq: phi}
x_{l+1,j}=\sum_{i=1}^{n_{in}^{l}} \phi_{l,i,j}(x_{l,i}),j=1,2,\dots,n_{out}^{l}
\end{aligned}
\end{equation} 
therefore, 
\begin{equation}
\begin{aligned}
\label{eq: Phi}
X_{l+1}=\underbrace{\left(\begin{array}{ccc}
\phi_{l, 1,1}(\cdot)  & \cdots & \phi_{l, 1, n_{in}^{l}}(\cdot) \\
\phi_{l, 2,1}(\cdot)  & \cdots & \phi_{l, 2, n_{in}^{l}}(\cdot) \\
\vdots &\ddots & \vdots \\
\phi_{l, n_{out}^{l}, 1}(\cdot) & \cdots & \phi_{l, n_{out}^{l}, n_{in}^{l}}(\cdot)
\end{array}\right)}_{\boldsymbol{\Phi}_l} X_l
\end{aligned}
\end{equation}
similarly, $X_{0}= H$ and $X_{\gamma+1}= \left\{ {y}_{1}^{\prime},{y}_{2}^{\prime},\dots,{y}_{c}^{\prime}\right \}$, where ${y}^{\prime}$ is the score of each classes.

For $\phi(\cdot)$, we have more details as follows:
\begin{equation}
\begin{aligned}
\label{eq:spline function}
\phi(x)=\mu \cdot b(x) +\omega \cdot spline(x)
\end{aligned}
\end{equation}
where $spline(x)$ is parametrized as a linear combination of B-splines as follows:
\begin{equation}
\begin{aligned}
\label{eq:B-spline}
spline(x)=\sum_{i}d_{i}\cdot B_{i}(x)
\end{aligned}
\end{equation} 
where $b(x)$ is Swish function \citep{ramachandran2017searching} and $d_{i}$ is trainable parametrize, $\mu$ and $\omega$ are redundant since it can be absorbed into $b(x)$ and $spline(x)$, $B_{i}(x)$ is the B-spline functions. Liu et al. \citep{liu2024kan} has already presented a comprehensive account of the parameter configuration and updating procedure. Consequently, this paper will not provide further elaboration.

To convert the model's output into predictive probabilities, we typically use a softmax function as follows:
\begin{equation}
\begin{aligned}
\label{eq:softmax}
{p}_{k}=\frac{e^{y_{k}^{\prime}}}{\sum_{i=0}^{c}e^{y_{i}^{\prime}}}
\end{aligned}
\end{equation}  

Therefore, the classification results of the corresponding time series at this moment are as follows:

\begin{equation}
\begin{aligned}
\label{eq: argmax}
{y}_{i}=\arg\max\limits_{i} p_{i}
\end{aligned}
\end{equation} 

Finally, we use the cross-entropy function \citep{rubinstein2004cross} as a loss function to measure the model's performance.

\section{Experiments}
\label{sec: Experiments}
In this section, we address the following questions by visualizing the model results:
\begin{enumerate}[\textbullet] 
    \item {Model accuracy}: We evaluate the generalization and classification performance of the models by comparing their results across various evaluation metrics.
    \item {Early predictive ability}: We assess the models' early prediction capabilities by conducting parameter sensitivity analyses, varying key parameters to examine their impact on predictive performance.
\end{enumerate}

\subsection{Dataset}
Data processing model the time series anomaly early prediction task as the series multi-classification task. Therefore, the model's classification performance reflects its generalization ability in early anomaly detection. The Natural Gas Pipeline Operation Detection (\textbf{NGPOD}) dataset \citep{pipedetect5}, \citep{0Leak}, \citep{zuo2022semi} consists of real-world data and is employed in this study. The model's effectiveness was validated using a widely recognized and publicly available benchmark benchmark: the Case Western Reserve University (\textbf{CWRU}) bearing dataset \citep{hendriks2022towards}.

The NGPOD dataset originates from real-world measurements gathered from natural gas transmission pipelines. These pipelines span a total length of 17.95 km and transport materials through four processing stations. Key variables analyzed include temperature, pressure, and flow rate signals. Temperature and pressure sensors were deployed at each station's inlet and outlet, enabling real-time monitoring. Notably, the first station has only an outlet, and the last station has only an inlet. Flow rate sensors were installed at the outlet of the first station and the inlet of the last station. Data collection occurred from 1 September 2020 to 25 December 2020, with sensors recording at 20-second intervals. In total, 364613 \textit{normal} and 12244 \textit{abnormal} samples were collected. Therefore, it is a highly imbalanced dataset.

The CWRU dataset is a well-established benchmark for time series anomaly detection. It was obtained using a custom-built bearing fault test rig that simulates various fault types. In addition to normal samples, faults were induced in diameter drive-end bearings of 7, 14, and 28 mils. The fault categories are subdivided into inner ring, rolling element, and outer ring faults. Each fault category contains 1198080 sampling points, allowing for classification across ten fault classes.

\subsection{Baselines}
\label{Baselines}
To assess our model's long-term data processing and generalization, we selected the following baselines:

\begin{enumerate}[1)] 
    \item \textbf{LSTM}: The Long Short-Term Memory (LSTM) network is a variant of recurrent neural networks that excels in sequential data by capturing long-term dependencies \citep{ding2018real}.
    
    \item \textbf{LSTM-KAN}: This model enhances the performance of LSTM by replacing its fully connected layer with a KAN, leveraging KAN's expressive power to model highly nonlinear relationships in complex time series data \citep{peng2024state}.
    
    \item \textbf{Transformer}: This self-attention mechanism-based architecture is effective in time series analysis for long-range dependency capture \citep{baseline-transformer}.

    \item \textbf{KAT}: It is a novel architecture that replaces Transformer's multilayer perceptron layers with KAN layers to enhance the expressiveness and performance of the model \citep{yang2024kolmogorov}.
    
    \item \textbf{TCN}: A convolutional neural network-based model tailored for time series data. TCN uses causal and dilated convolutions for effective temporal dependency modeling \citep{baseline-tcn}.
\end{enumerate}

\subsection{Evaluation Metrics}
\label{sec: metrics}

The paper adopts a macro evaluation framework to ensure equal weighting for each class in a sample imbalance. We use multiple metrics to compare against baselines and evaluate the model from various perspectives. Accuracy \citep{eledkawy2024precision} and Kappa \citep{sim2005kappa} reflect the overall model performance, while Negative Predictive Value (NPV) \citep{steinberg2009sample} and Precision \citep{liu2019time} assess the prediction accuracy for negative and positive classes. Specificity \citep{liu2019time} and Recall \citep{ge2024enhanced} evaluate the model's ability to identify negative and positive classes, respectively. F1-Score \citep{liu2019time} and G-measure \citep{lu2021using} offer balanced metrics by combining Precision with Recall and Sensitivity with Specificity. Finally, Matthews Correlation Coefficient (MCC) \citep{chicco2020advantages} assesses model consistency, and Area Under the Negative Predictive Value-Recall Curve (AUNP) \citep{benzekry2021machine} captures the relationship between NPV and Recall across varying thresholds.

\subsection{Experimental Setup}

\begin{figure}[t]
\centering 
\includegraphics[width=3.45in]{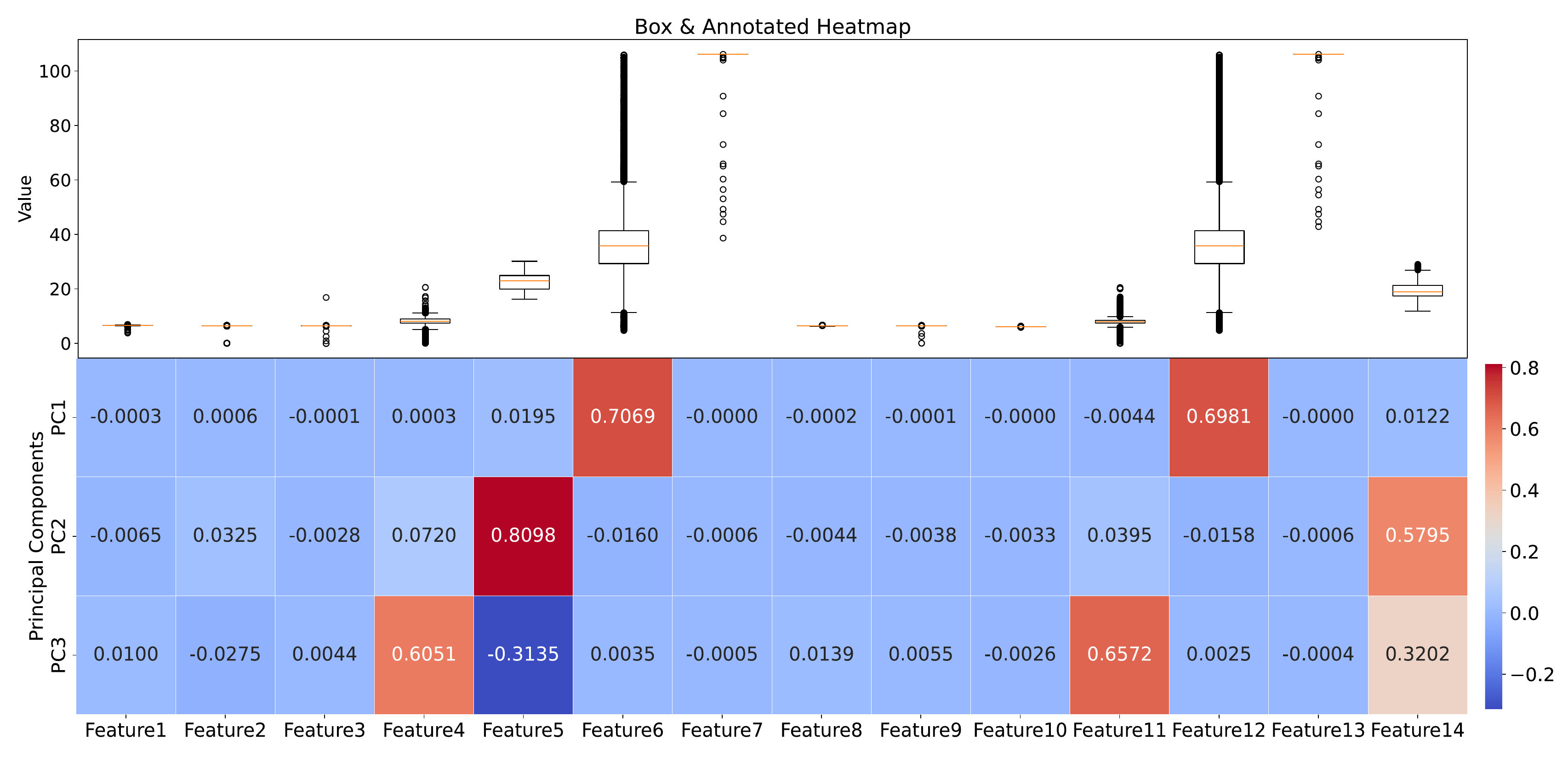} 
\caption{A box-and-line plot is employed to describe the original data distribution. At the same time, a heat map is utilized to illustrate the contribution of each feature in the original data, which PCA has processed.} 
\label{fig: pca} 
\end{figure}

\begin{table}[t]
\centering
\caption{Number of hidden layers per model and total number of parameters.}
\label{tab:model parameteter}
\begin{tabular}{ccc}
\hline\hline
\textbf{Model}       & \textbf{Hidden layer} & \textbf{Parameters} \\ \hline
LSTM         & [128, 64]      & 153987     \\
LSTM-KAN     & [128, 64]      & 153184     \\
Transformer  & [32]           & 404681     \\
KAT          & [32]           & 404572     \\
TCN          & [32, 64, 128]  & 126179     \\
MKTCN        & [32, 64, 128]  & 119632     \\ \hline\hline
\end{tabular}
\end{table}

The paper employs an overlapping sliding window approach to extract a series of sub-sequences from the datasets. We define the sliding window size as $\omega=50$ and the sliding step as $s=1$. The data were randomly partitioned into the training set (70\%), validation set (20\%), and test set (10\%). After applying Principal Component Analysis (PCA) for dimensionality reduction, the retained features account for a minimum contribution rate of 95\%.

For each model described in Section \ref{Baselines}, the model set the batch size to 64, the dropout to 0.5, the kernal size to 3, the learning rate to 0.001, and the Adam as optimizer. The number of epochs for the NGPOD dataset was 10, while the number for the CWRU dataset was 200. Furthermore, the number of multi-head attentions for the Transformer and KAT models was set to 3, the kernel size for the TCN and MKTCN models was set to 3, and the grid size for the LSTM-KAN and MKTCN models was set to 5. The number of hidden layers and parameters for each model is presented in Table \ref{tab:model parameteter}.

\subsection{Models Comparison}
\label{Models Comparison}

\begin{table*}[t]
\centering
\caption{The quantitative results of the six models when $N$ = 200 in two benchmarks. The best results are shown in \textbf{bold}, and the second best results are \underline{underlined}.}
\setlength{\tabcolsep}{0.5mm}
\label{tab: model comparison}
\begin{tabular}{cccccccccccc}
\hline \hline
\textbf{Dataset}       & \textbf{Model} & \textbf{Accuracy} & \textbf{NPV}    & \textbf{Precision} & \textbf{Specificity} & \textbf{Recall} & \textbf{F1-Score} & \textbf{AUNP}   & \textbf{Kappa}  & \textbf{MCC}    & \textbf{G-Measure}    \\ \hline
\multirow{6}{*}{NGPOD} & LSTM           & 0.9769            & 0.9406          & 0.7907             & \underline{0.8978}         & \underline{0.6979}    & \underline{0.7379}      & \underline{0.8467}    & \underline{0.7295}    & \underline{0.6594}    & 0.7411                \\
                       & LSTM-KAN       & \underline{0.9835}      & \underline{0.9426}    & \underline{0.8326}       & 0.8865               & 0.6667          & 0.7249            & 0.8297          & 0.7136          & 0.6500          & 0.7368                \\
                       & Transformer    & 0.9804            & 0.9421          & 0.7507             & 0.7785               & 0.4625          & 0.5068            & 0.6677          & 0.4727          & 0.3647          & 0.5207                \\
                       & KAT            & 0.9795            & 0.9207          & 0.7548             & 0.7774               & 0.4614          & 0.5015            & 0.6661          & 0.4591          & 0.5037 & \underline{0.7688} \\
                       & TCN            & 0.9820            & 0.9412          & 0.7367             & 0.8908               & 0.6880          & 0.7074            & 0.8362          & 0.7068          & 0.6248          & 0.7099                \\
                       & \textbf{MKTCN} & \textbf{0.9840}   & \textbf{0.9661} & \textbf{0.8648}    & \textbf{0.9037}      & \textbf{0.7592} & \textbf{0.8058}   & \textbf{0.8555} & \textbf{0.7739} & \textbf{0.7421} & \textbf{0.8089}       \\ \hline
\multirow{6}{*}{CWRU}  & LSTM           & 0.9839            & 0.9911          & 0.9242             & 0.9911               & 0.9230          & 0.9212            & 0.9552          & 0.9105          & 0.9137          & 0.9224                \\
                       & LSTM-KAN       & 0.9960      & 0.9978    & 0.9819       & 0.9978               & 0.9814          & 0.9813      & 0.9888          & 0.9776          & 0.9792          & 0.9815                \\
                       & Transformer    & 0.9884            & 0.9936          & 0.9507             & 0.9935               & 0.9439          & 0.9477            & 0.9673          & 0.9354          & 0.9357 & 0.9494       \\
                       & KAT            & \underline{0.9991}            & \underline{0.9995}          & \underline{0.9960}             & \underline{0.9995}         & \underline{0.9959}    & \underline{0.9959}            & \underline{0.9975}    & \underline{0.9950}    & \underline{0.9954}    & \underline{0.9959}          \\
                       & TCN            & 0.9880            & 0.9933          & 0.9454             & 0.9933               & 0.9426          & 0.9430            & 0.9663          & 0.9329          & 0.9369          & 0.9435                \\
                       & \textbf{MKTCN} & \textbf{0.9996}   & \textbf{0.9998} & \textbf{0.9981}    & \textbf{0.9998}      & \textbf{0.9980} & \textbf{0.9980}   & \textbf{0.9987} & \textbf{0.9975} & \textbf{0.9978} & \textbf{0.9980}       \\ \hline \hline
\end{tabular}
\end{table*}

To evaluate the generalization and classification performance of the models, we compare their results across various evaluation metrics. Firstly, we transformed the multi-dimensional time series into sequential data using the data processing step. In Fig.\ref{fig: pca}, we use a box-and-line plot (upper part) to describe the original data distribution and a heat map (lower part) to describe the contribution of each feature in the original data processed by PCA. It can be observed that features 6 and 12 contribute the most to principal component 1 (74.56\% contribution), features 5 and 14 contribute the most to principal component 2 (15.85\% contribution), and features 4 and 11 contribute the most to principal component 3 (5.29\% contribution). Consequently, we selected the top three principal components. The application of PCA enables the extraction of the data's critical features while simultaneously enhancing the predictive performance of the models (all baselines and MKTCN).

Then, we use the metrics outlined to evaluate the models described on the NGPOD and CWRU datasets. Table \ref{tab: model comparison} presents the quantitative results for the six models when $N = 200$. Comparative results of different models are as follows:

\begin{enumerate}[\textbullet] 
\item For the NGPOD dataset, the MKTCN model performs best. This demonstrates that the dilated causal convolution module effectively handles data with long-range dependencies. Specifically, MKTCN resulted in improvements in Accuracy, NPV, Precision, Specificity, Recall, F1-score, AUNP, Kappa, G-measure, and MCC, with values increasing by 0.0005, 0.0235, 0.0322, 0.0059, 0.0613, 0.0679, 0.0088, 0.0444, 0.0827 and 0.0401, respectively.
\item For the CWRU dataset, the classification ability across all metrics follows the order: MKTCN $>$ KAT $>$ LSTM-KAN $>$ Transformer $>$ TCN $>$ LSTM. The MKTCN model excels in all metrics, confirming the efficacy and precision of its design, particularly for multi-classification tasks.
\end{enumerate}

The main finding of the \textbf{ablation study} is as follows:

\begin{enumerate}[\textbullet] 
\item Using the NGPOD dataset as an example, the MKTCN model can generalize more than the TCN model. Specifically, MKTCN improved all metrics, increasing the metric values by 0.0020, 0.0249, 0.1281, 0.0129, 0.0712, 0.0984, 0.0193, 0.0671, 0.1173, and 0.0990, respectively.
\end{enumerate}

\begin{figure*}[t]
\centering 
\includegraphics[width=6.3in]{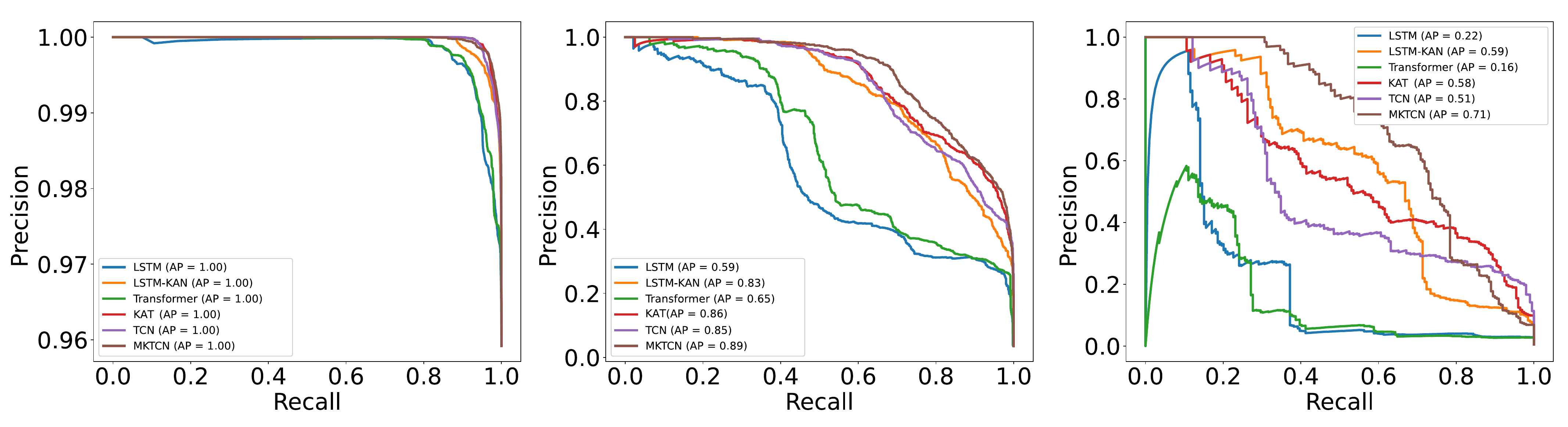} 
\caption{Precision-Recall curves in the NGPOD dataset. (a) \textit{Normal} class. (b) \textit{Abnormal} class. (c) \textit{Doubtful} class. } 
\label{fig:PR} 
\end{figure*}

Meanwhile, we use the Precision-Recall (PR) curves in Fig. \ref{fig:PR} to represent model performance visually. Average Precision (AP) assesses a model's precision across different recall levels, representing the area under the PR curve. Fig. \ref{fig:PR}(a) demonstrates that classification outcomes for the normal class are generally comparable across all models. However, as shown in Fig. \ref{fig:PR}(b) and Fig. \ref{fig:PR}(c), significant discrepancies appear in the abnormal and doubtful classes, mainly due to sample imbalance. Specifically, for the \textit{Normal} class, the AP of all models is 1. For the \textit{Abnormal} class, the effects of the models are in the order of MKTCN (0.89) $>$ KAT (0.86) $>$ TCN (0.85) $>$ LSTM-KAN (0.83) $>$ Transformer (0.65) $>$ LSTM (0.59). For the \textit{Doubtful} class, the order of the models' effects is as follows: MKTCN (0.71) $>$ LSTM-KAN (0.59) $>$ KAT (0.58) $>$ TCN (0.51) $>$ LSTM (0.22) $>$ Transformer (0.16). The MKTCN model performs better in these classes, reflecting its enhanced generalization capability.

\begin{figure}[t]
\centering 
\includegraphics[width=2.2in]{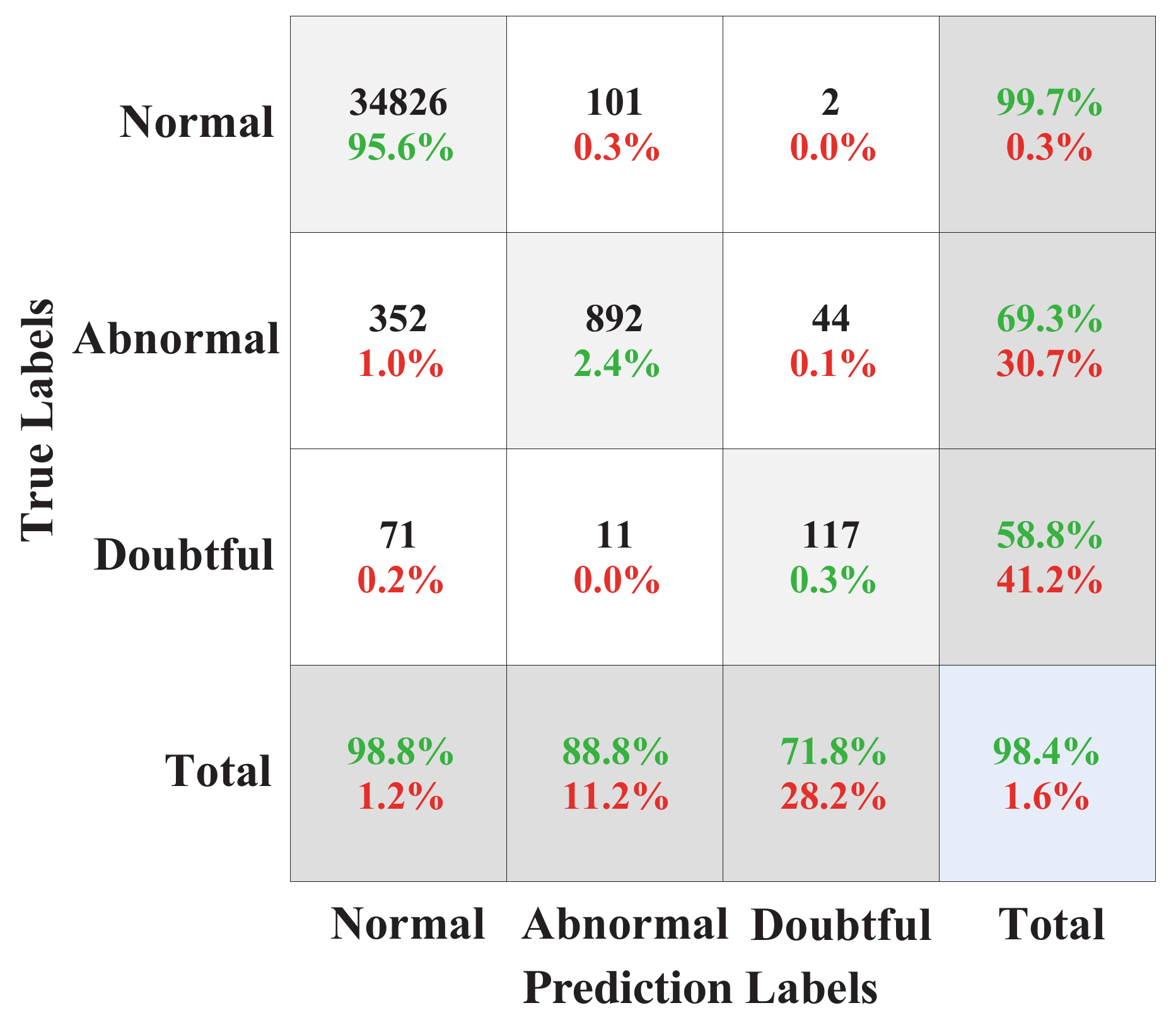} 
\caption{Confusion matrix for the MKTCN model on NGPOD dataset. Proportion is expressed as a percentage of the total. The correct statistic is indicated by green, while the incorrect statistic is indicated by red.} 
\label{Confusion matrix for NGPOD} 
\end{figure}

\begin{figure}[t]
\centering 
\includegraphics[width=3.45in]{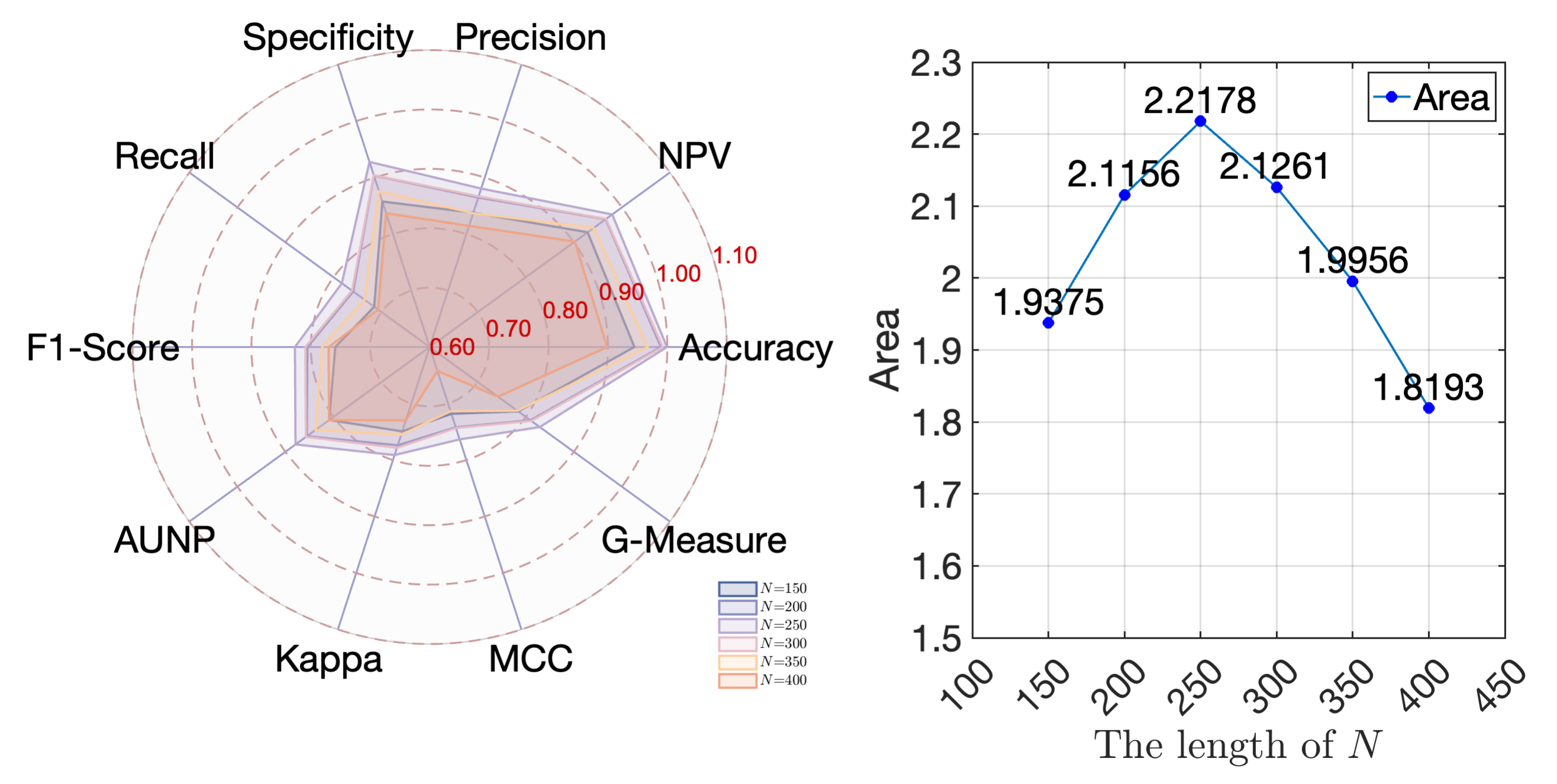} 
\caption{Comparison of metrics using radar charts for the NGPOD dataset. (a) Radar charts are used to depict the values of different evaluation metrics at different $N$'s. (b) Each point on the line charts represents the area enclosed by the radar plot corresponding to the different $N$.} 
\label{radar for NGPOD} 
\end{figure}

Fig. \ref{Confusion matrix for NGPOD} shows the confusion matrix of the MKTCN model for the multi-classification task on the NGPOD dataset. The MKTCN model achieves an overall accuracy of 98.40\%. For the \textit{normal} class, it attains an accuracy of 98.8\%, accurately classifying all normal samples. The \textit{abnormal} and \textit{doubtful} classes achieve accuracies of 88.8\% and 71.8\%, respectively. This indicates that while the model excels in distinguishing the three categories, significant confusion remains between the \textit{abnormal} and \textit{doubtful} categories, likely due to the limited sample size. Despite this challenge, the MKTCN model still substantially improves over other models.

In conclusion, the MKTCN model outperforms the other four models on both datasets, showcasing its superior generalization and time-series forecasting capabilities. It can predict pipeline leakage events up to 4000 seconds in advance ($N=200$).

\subsection{Different Time Lengths for Early Leaks Prediction}
To assess the early prediction capabilities of the models, we will conduct sensitivity analyses on key parameters to examine their impact on predictive performance. In Section \ref{Models Comparison}, we established the effectiveness of the MKTCN model through comparative analysis with other approaches. Furthermore, we demonstrated that the MKTCN model can predict leaks up to 4,000 seconds in advance. In this section, we evaluate the predictive performance of the MKTCN model by varying the parameter $N$, which determines how far in advance the model can capture early warning signals before a leak occurs. Table \ref{tab: different l} shows the details of the data set under different $N$. Radar plots are employed to illustrate the performance of the MKTCN model at different values of $N$, as shown in Fig.\ref{radar for NGPOD} (a). Subsequently, Fig.\ref{radar for NGPOD} (b) presents the area enclosed by the radar plots corresponding to each value of $N$.

\begin{table}[t]
\centering
\caption{Detailed information on NGPOD datasets under different time lengths  $N$ for early Leaks prediction.}
\label{tab: different l}
\begin{tabular}{cccc}
\hline\hline
\textbf{$N$}               & \textbf{\textit{Normal}}  & \textbf{\textit{Doubtful}} & \textbf{\textit{Abnormal}} \\ \hline
150          & 351469          & 900               & 12244            \\
{200} & {351180} & {1189}    & {12244}   \\
250          & 350930          & 1439             & 12244            \\
{300} & {350680} & {1689}    & {12244}   \\
350          & 350430          & 1939             & 12244            \\
400          & 350180          & 2189             & 12244            \\ \hline\hline
\end{tabular}
\end{table}

Since the evaluation metrics described in Section \ref{sec: metrics} positively correlate with model performance, we consider them collectively. MKTCN is optimal across a range of $N$ compared to other baselines. The results indicate that the enclosed area of the radar plots reaches its maximum when $N$ is set to 250. This suggests that the MKTCN model achieves optimal accuracy when predicting 5000 seconds ($N=250$) before a leak occurs. If $N$ is smaller than 250, the model may need help learning sufficient samples due to the limited number of $\textit{doubtful}$ samples, leading to poorer performance. On the other hand, when $N$ exceeds 250, it is possible that the data no longer effectively reflects early leak characteristics, resulting in the model learning irrelevant features, thus degrading prediction accuracy.

\section{Conclusion}
\label{sec: Conclusion}
This paper addresses a critical gap in the early prediction of pipeline leaks by modeling the time series prediction task as a sequential multi-class classification task. We introduce a novel model, the MKTCN, which integrates the robust long-term sequence modeling capabilities of dilated convolution with the strong generalization abilities of KAN.

\bibliographystyle{unsrtnat}
\bibliography{references}  %%% Uncomment this line and comment out the ``thebibliography'' section below to use the external .bib file (using bibtex) .





% \end{thebibliography}

\end{document}